\documentclass[%
 preprint,
 aps,
 pre,
 amsmath,amssymb
]{revtex4-2}

\usepackage{graphicx}
\usepackage{dcolumn}
\usepackage{esvect}
\usepackage{bm}
\usepackage{algorithm}
\usepackage{algpseudocode}
\usepackage{soul}

\usepackage[percent]{overpic}

\usepackage[dvipsnames]{xcolor}
\definecolor{roie}{rgb}{0.858, 0.188, 0.478}

\begin{document}


\title{Harnessing intuitive local evolution rules for physical learning}

\author{
  Roie Ezraty$^{1}$, 
  Menachem Stern$^{2}$, 
  Shmuel M. Rubinstein$^{1}$\\
  \small $^1$Racah Institute of Physics, Hebrew University of Jerusalem, Jerusalem 9190401, Israel\\
  \small $^2$AMOLF, Science Park 104, 1098 XG Amsterdam, The Netherlands\\
  \small \texttt{roie.ezraty@mail.huji.ac.il}
}

\date{\today}

\begin{abstract}

Machine Learning, however popular and accessible, is computationally intensive and highly power-consuming, prompting interest in alternative physical implementations of learning tasks. We introduce a training scheme for physical systems that minimize power dissipation in which only boundary parameters (i.e. inputs and outputs) are externally controlled. 
Using this scheme, these Boundary-Enabled Adaptive State Tuning Systems (BEASTS) learn by exploiting local physical rules. Our scheme, BEASTAL (BEAST-Adaline), is the closest analog of the Adaline algorithm for such systems. We demonstrate this autonomous learning in silico for regression and classification tasks.
Our approach advances previous physical learning schemes by using intuitive, local evolution rules without requiring large-scale memory or complex internal architectures. BEASTAL can perform any linear task, achieving best performance when the local evolution rule is non-linear.

\end{abstract}

\maketitle

\section{Introduction}\label{sec:Introduction}

    Artificial Intelligence and Machine Learning revolutionized our lives in the last decades; from performing elaborate mathematical computations to playing chess at highest levels \cite{baktash2023gpt4reviewadvancementsopportunities} and predict protein structure \cite{senior2020improved}. 
    Artificial-Neural-Networks (ANNs) are at the forefront of these advancements; Inspired by the structure of the brain \cite{burkitt2004spike}, ANNs have the remarkable ability to learn multiple tasks and solve difficult inverse-design problems \cite{lecun2015deep}. 
    The outstanding versatility of the ANNs is achieved when the number of tunable Degrees of Freedom (DOFs, i.e., weights) is very large. The training process typically relies on gradient descent (GD), which updates all the tunable DOFs simultaneously by following the gradient of a global loss function \cite{rumelhart1986learning}. This process is performed in a highly multidimensional space and the training process requires many billions of computations, making it highly power consuming \cite{luccioni2023estimating, mytton2022sources}. 
    Even an efficient online GD implementation such as the Adaline algorithm \cite{widrow1988adaptive}, which involves single layer, fully connected linear networks with linear activations, is cumbersome and requires access to all DOFs simultaneously. 
    Furthermore, ANNs rely on centralized computing architectures under a global processor, and as their size increases, training-time increases significantly, leading to substantial slowdowns \cite{bahri2024explaining}.
    
    ANNs have surpassed our abilities in numerous tasks, but nevertheless natural systems still hold several advantages. A brain learns very quickly \cite{zucker2002short, kanai2005perceptual, FRENCH1999128} and efficiently \cite{sengupta2014power}, and appears to capitalize on local rules for synaptic plasticity. Moreover, it is versatile for a variety of tasks and is robust against damage \cite{mcgovern2019hemispherectomy}. Other natural systems show similar abilities: amoebas can solve the traveling-salesman problem \cite{zhu2013amoeba}, artificial physical systems such as thin sheets can be trained to self-fold into desired shapes \cite{arinze2023learning}, memristor arrays can solve linear classification and regression \cite{sun2020one, sun2019solving} and laser light can find the ground state of an $XY$ spin lattice model \cite{Davidson2020PhysRevResearch}. 
    
    Harnessing physical systems for learning tasks \cite{stern2023learning, lopez2023self, martinez2024fluidic,  wright2022deep}, i.e., physical learning, has lately received much attention, especially in power minimizing systems which self optimize their power use \cite{vadlamani2020physics}. 
    The advantages extend beyond power efficiency and speed, as    physical systems can also interact with and affect their physical environment; they could integrate calculation and functionality in a single medium. 
    Implementing GD algorithms on physical learning systems could hence be of great benefit, since those can accommodate learning with minimal external intervention.
    
    It was recently shown that physical materials that minimize elastic energy and whose internal parameters change due to known local rules can be trained to perform a desired functionality. This training procedure, termed Directed Aging \cite{pashine2019directed, hexner2020effect}, where the outputs are forced to maintain a desired state, was demonstrated in materials such as mechanical \cite{stern2020continual}, spring and spin networks \cite{gold2019self, kedia2019drive}. The scheme capitalizes on local evolution rules without direct knowledge of the internal DOFs. It is, however, limited in the scope of tasks that can be performed; there is no loss function being optimized, so specific input-output relations cannot be directly encoded, as in supervised learning. 

    More versatile training schemes that approximate GD and capitalize on local physical rules in systems that minimize power dissipation are the Contrastive Learning schemes. Examples of which include coupled learning \cite{stern2021supervised, anisetti2023learning, xie2003equivalence}, contrastive Hebbian learning \cite{movellan1991contrastive} and equilibrium propagation \cite{scellier2017equilibrium, scellier2021deeplearningtheoryneural}, which are inspired by neuroscience and learning theory \cite{hinton2002training, ackley1985learning}. These can solve tasks more stringent than those demonstrated with Directed Aging \cite{dillavou2022demonstration, altman2024experimental, falk2025temporal}. 
    In Contrastive Learning, supervised training is done by presenting the system with specific examples of an abstract desired function, namely: linear regression or classification. The system lowers a loss based on the examples by evolving its internal parameters following local evolution rules \cite{stern2024physical}. As the examples are presented and the system evolves, it eventually learns not only the presented examples but also generalizes.
    This is a decentralized, physics-based learning that can be very efficient in terms of power dissipation \cite{stern2024training}. However, it requires information from two distinct physical states, usually in the form of external memory or twin couples of the same system. Accessing or computing each internal parameter at all times is still a standing difficulty in the field of physical learning networks. 
    Both Directed Aging and Contrastive Learning train systems where the only parameters directly accessible are the inputs and outputs at the system boundaries, so they are termed here \emph{Boundary-Enabled Adaptive State Tuning Systems} (BEASTS). The pending question is: can GD be implemented on BEASTS whose internal parameters evolve due to intuitive local physical dynamics, rather than abstract mathematical learning rules? 

    In this work, we develop a physical learning scheme for training BEASTS where internal parameters change due to natural and intuitive physical rules, e.g. fluidic resistances that adjust due to local hydrostatic pressure drops.
    The perspective here is the opposite of that in Contrastive Learning: instead of deriving a local rule that yields a material with learning capabilities if put under the right boundary conditions and then intricately building such a material, we start from materials that have intuitive local rules and implement a general procedure that trains them using solely boundary conditions.
    This scheme is applicable to systems that minimize power dissipation, so they are modeled here as fluidic resistor networks whose resistances (or learning DOFs \cite{dillavou2022demonstration}) change due to local learning rules. 
    This is the closest version of the efficient Adaline algorithm applicable to BEASTS, so is termed the BEASTAL (BEAST Adaline-Like) scheme. 
    With this minimal external supervision, we train BEASTS with a plethora of local rules to perform linear regression and classification tasks. We emphasize the efficiency of the scheme, while shedding light on the importance of non-linearities in the system for improved learning.   

\section{Methods}\label{sec:Methods}


    In this work, a general family of resistor networks is examined. We consider a fluidic system where resistors are tubes through which water flows due to hydrostatic pressure differences $\Delta p_{ij}=p_i-p_j$. Fluidic resistors $R_{ij}$, are network edges connecting nodes $i$ and $j$ and $p_i$ is the pressure at node $i$ (see Table \ref{tab:nomenclature}).
    Input pressures $\vec{x}$ are given at the input nodes, and output pressures $\vec{y}$ are measured at the output nodes. Given the inputs, the pressure on the rest of the nodes, and specifically on the outputs, is dictated by minimization of total power dissipation, defined as
    
    \begin{equation}\label{eq:power_dissipation}
        \Pi=\sum \frac{\Delta p_{ij}^2}{R_{ij}} \ ,
    \end{equation}
    where the summation is over all edges. Power minimization is a deterministic physical process.    
    
    The system can be in one of two modalities: a) Measurement modality, where resistances are constant, and outputs are measured given the inputs, and b) Update modality where the resistances change locally, according to an evolution rule, for example:

    \begin{equation}\label{eq:Rdot_afo_deltap}
        \dot{R}_{ij}\left(t\right)=\gamma \Delta p^!_{ij}\left(t\right) \ ,
    \end{equation}
    where $\gamma$ is a proportionality constant, which could be positive or negative, and the superscripted exclamation mark $^{\,!}$ symbolizes that the system is in the Update modality. We later consider different local rules than equation \ref{eq:Rdot_afo_deltap}.  
    Experimentally, the local evolution rule (equation \ref{eq:Rdot_afo_deltap}) can be implemented in materials or systems where resistors evolve based on pressure differences or flow; for example, if the resistors are fluidic tubes that clog or unclog due to pressure difference or flow across them, or shear-thickening fluids where the hydraulic resistance varies with flow rate. 
    A physically justified rule for the evolution of rest lengths in a mechanical spring network is introduced in \cite{hexner2020periodic} which is similar to equation \ref{eq:Rdot_afo_deltap}.
     
    In classical ANN methods, the learning DOFs are weights, and they are all directly manipulated during training. In physical systems, however, the user has direct access only to inputs and outputs, and physical laws dictate the evolution of the internal DOFs ($R_{ij}$ in BEAST systems). Typically, the number of inputs and outputs is much smaller than the number of learning DOFs, and we would like to capitalize on this fact. In the following sub-section, we devise a training scheme for BEASTS that obey equations \ref{eq:power_dissipation} and \ref{eq:Rdot_afo_deltap}. It smartly uses boundary conditions to evolve the internal DOFs of so as to minimize a loss function.

    \begin{table}[b]
    \caption{\label{tab:nomenclature}
    Relevant Nomenclature for the training scheme.
    }
    \begin{ruledtabular}
    \begin{tabular}{lcdr}
    \textrm{Notation}&
    \textrm{Meaning}\\
    \colrule
        $p$ & node pressures in all the network\\ 
        $\vec{x}$  & input pressures, dimension $\#_{\text{Inputs}}$\\
        $\vec{y}$  & output pressures, dimension $\#_{\text{Outputs}}$\\
        $R$  & Resistances\\
        $\vec{\mathcal{L}}$ & Loss, dimension $\#_{\text{Outputs}}$\\
        $\alpha$  & learning rate, scalar\\
        $\gamma$  & resistance-pressure ratio coefficient, scalar\\
        $\boxed{}^{!}$  & Update modality value, used for resistances update\\
        $\widehat{\boxed{}}$  & desired value of the task\\
        $t$  & training step $\#$
    \end{tabular}
    \end{ruledtabular}
    \end{table}  

\subsection{Training the network}\label{sec:training}

    This work examines networks where each input node is connected to each output node and all are connected to a ground node with a fixed pressure of $0$, so there are $\left(\#_{\text{Inputs}}+1\right) \times\#_{\text{Outputs}}$ edges in total. 
    Adding hidden layers will not improve the performance of linear networks (Appendix \ref{app:hidden_layers}). It can only hinder network performance and we stick to this simple structure.
    
    In common online training methods for ANNs, a single random sample $\vec{x}$ is drawn from a dataset at each training step, outputs $\vec{y}=\vec{f}_W\left(\vec{x}\right)$ are measured, where $W$ are the weights, and a loss vector is calculated:
    
    \begin{equation}\label{eq:loss}    
        \vec{\mathcal{L}}\left(t\right)=\widehat{\vec{y}}\left(t\right)-\vec{y}\left(t\right) \ .
    \end{equation}
    All the weights are incrementally changed in order to minimize the Mean Squared Error (MSE) $\overline{\|\vec{\mathcal{L}}\|^2}$. Classically, the change is calculated using backpropagation or other numerical algorithms. This concludes a single training step. The process is repeated until the loss is lowered sufficiently. 
    
    Here, at each training step the system starts with a Measurement $\vec{y}=\vec{f}_R\left(\vec{x}\right)$,
    during which internal DOFs, i.e. the resistances, do not change.  
    The system then switches to the Update modality where inputs as well as outputs are imposed, and resistances evolve according to equation \ref{eq:Rdot_afo_deltap}. The supervisor has limited access and limited information of the state of internal DOFs. The challenge is therefore to impose inputs and outputs that will change the resistances in a way that lowers the MSE loss. 
    An approximate Adaline implementation on resistor networks requires changing the resistances of all edges such that $\dot{R}_{ij}\propto\left(y_{i} - x_{j}\right) \left(\hat{y}_{i}-y_{i}\right)$ (see Appendix \ref{app:derivation_Adaline_like}), but since BEASTS only have $\#_{\text{Inputs}}+\#_{\text{Outputs}}$ controls, desired resistance changes cannot all be satisfied simultaneously.
    We propose an Adaline-like scheme that successfully trains BEASTS, BEASTAL, where the Update inputs and outputs are:

    \begin{equation}\label{eq:Adaline-like}
    \left[\begin{array}{c}
    \vec{x}^{!}\\
    \vec{y}^{!}
        \end{array}\right]=\frac{\alpha}{\gamma}U^{\dagger}\overrightarrow{\left(y_{i} - x_{j}\right) \left(\hat{y}_{i}-y_{i}\right)}
    \end{equation}
    with 
    $\alpha$ the learning rate constant, $i$ goes over all outputs and $j$ over all inputs+ground and $U^\dagger$ is the pseudo-inverse of the incidence matrix (Appendix \ref{app:incidence_mat}) which contains the relation between $\#_{\text{Inputs}}+\#_{\text{Outputs}} +1$ nodes and $\left(\#_{\text{Inputs}}+1\right) \times\#_{\text{Outputs}}$ edges (thorough derivation in Appendix \ref{app:derivation_Adaline_like}). $\overrightarrow{\left(y_{i} - x_{j}\right) \left(\hat{y}_{i}-y_{i}\right)}$ has a value for every $i$ of the outputs and every $j$ of the inputs+ground so has the dimensions of $\#_{\text{Inputs}}+\#_{\text{Outputs}} +1$ which is the number of edges. 
    
    Equation \ref{eq:Adaline-like} yields the closest version of Adaline that can be implemented on BEASTS. There, $\dot{R}_{i}\propto\frac{1}{\#_{edges}}\left(\left(\#_{inputs}+\#_{outputs}\right)\Delta p_{i}+\sum_{k\neq i}\left(\varepsilon_k\Delta p_{k}\right)\right)$ with $\varepsilon_k\leq\#_{inputs}$, so the major contribution to the resistance change on edge $i$ is $\Delta p_i$ for all $i$. Network evolution under these dynamics is mostly down the gradient of the loss (Appendix \ref{app:comparison_GD}). Importantly, the Update modality values depend only on network topology and measured quantities, and are not mere samples from the dataset $x_i,\,y_j$.
    For clarity, the general training scheme is illustrated as a pseudocode in the Training Scheme Algorithm.


    \begin{center}
    \fbox{\parbox{0.95\linewidth}{
    \textbf{Training Scheme Algorithm} \\[0.5em]
    \textbf{for} $t \in \#_{\text{iterations}}$ \textbf{do} \\
    \hspace*{1em} Sample $\vec{x}$ \\
    \hspace*{1em} $\hat{\vec{y}} = M\vec{x}$ \hfill  desired outputs \\
    \hspace*{1em} Measure $\vec{y}$ under minimization of $\Pi$ \\
    \hspace*{1em} $\vec{\mathcal{L}}(t) = \hat{\vec{y}}(t) - \vec{y}(t)$ \hfill losses \\
    \hspace*{1em} $\left[\begin{array}{c}
                 \vec{x}^{!}\\
                 \vec{y}^{!}
                 \end{array}\right]=\frac{\alpha}{\gamma}U^\dagger
                 \overrightarrow{ \left(y_{i} - x_{j}\right) \left(\hat{y}_{i}-y_{i}\right)}$ \hfill Update \\
    \hspace*{1em} $\Delta p^{\,!}_{ij}$ under minimization of $\Pi$ \\
    \hspace*{1em} $R_{ij}\left(t\right)=R_{ij}\left(t-1\right) + \gamma \Delta p^!_{ij}\left(t\right)$ \hfill or other evolution rule \\
    \textbf{end for}
    }}\label{training_scheme_algorithm}
    \end{center}

    In the next section we show that the learning dynamics follow gradient-descent. Here
    we give a brief intuition by examining first the case where the desired output at node $i$ is lower than the measured $\hat{y}_i<y_i$: 
    Note that $y_i$ is mostly lower than $x_j$ due to the ground node, therefore according to equation \ref{eq:Adaline-like} resistances $R_{ij}$ increase. The amount of resistance increase is proportional also to the measured pressure difference across it. Higher resistances can accommodate larger pressure drops so in the next training step, the measured values will be lower, closer to desired ones. A similar argument applies when the desired outputs are initially lower.

\section{Simulation Results}\label{sec:results}

    Using numerical simulations of fluidic resistor networks at steady state, we first show that BEASTS can perform 1) linear regression and 2) classification tasks. We then examine the performance as task complexity, and therefore network size, scales up, showing that a non-linearity in the evolution rule for resistances greatly improves learning. Lastly, in the conclusions, we study various evolution rules in which edge-resistance changes according to different local forcings, and successfully train them all using the AL scheme.
    
\subsection{Linear Regression - Linear evolution rule}\label{sec:linear_regression}

    \begin{figure*}[ht]
    \centerline{   
    \includegraphics[width=\textwidth,height=\textheight,keepaspectratio]{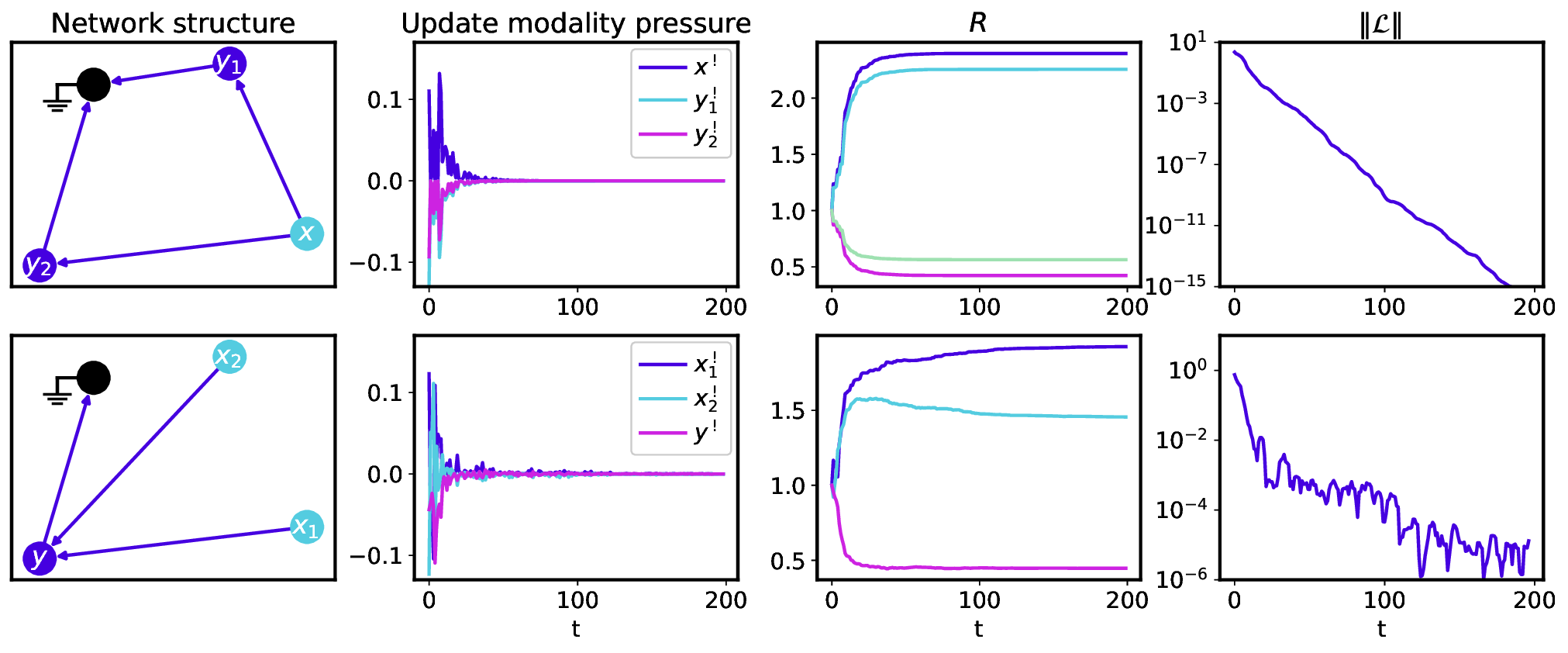}
    }
    \caption{Network performance on two regression tasks: A network with $2$ input, $1$ output and $1$ ground nodes trained for the task $\left[\begin{array}{c}y_1\\y_2\end{array}\right]=\left[\begin{array}{c}0.15\\0.2\end{array}\right]x$ in the upper panels and a network with $1$ input $2$ output and $1$ ground node trained for the task $y=\left[\begin{array}{cc}0.15 & 0.2\end{array}\right]\left[\begin{array}{c} x_{1}\\x_{2}\end{array}\right]$ in the bottom panels. Left to right - the network structure with lines denoting edges; input nodes in cyan, output nodes in purple and ground in black (arrows point from inputs to outputs and from outputs to ground), input and output pressures during the Update modality, resistances and normalized MSE loss, all as function of training step $t$. The loss is normalized at every time step by dividing equation \ref{eq:loss} by the mean squared value of the desired outputs. In both cases, the MSE loss decreases to arbitrarily low values}
    \label{fig:performance_2_networks}
    \end{figure*}

    A common task for physical learning systems \cite{altman2024experimental, dillavou2022demonstration}, as well as artificial neural networks \cite{bishop2006pattern, tibshirani1996regression}, is linear regression, as in

    \begin{equation}\label{eq:task}
        \widehat{\vec{y}}=M\vec{x} \ ,
    \end{equation}
    where $M$ is a matrix with dimensions $\#_{\text{Inputs}} \times\#_{\text{Outputs}}$. BEASTS are limited in the breadth of regression tasks they can learn, since their output values cannot exceed the input values. Namely, the sum over each row of the task matrix $M$ can't exceed $1$, but amplifying the outputs of the network by a constant negates this pitfall.
    
    Fluidic resistor networks with a simple fully connected structure learn low-dimensional regression tasks when trained using BEASTAL, as shown in figure \ref{fig:performance_2_networks}. Low dimensional regression with a small training set and no test set had been related to problems of allostery in proteins and flow networks \cite{rocks2017designing, stern2021supervised, ribeiro2016chemical}. BEASTAL therefore successfully trains for allostery tasks. Moreover, it is applicable to more complicated regression tasks, i.e., numerous inputs and outputs, and to various task matrices $M$, as shown in the left panel of figure \ref{fig:log_loss_afo_inputs_outputs} (see further explanation in Appendix \ref{app:net_structure}). 
    Even though performance decreases as task complexity increases, BEASTAL outperforms a non-iterative analytical approach (Appendix \ref{app:pseudo_inverse}).
    Moreover, in the next subsection we show that a non linear evolution rule greatly improves performance. 

    \begin{figure}[ht]
    \centerline{
    \includegraphics[width=\columnwidth]{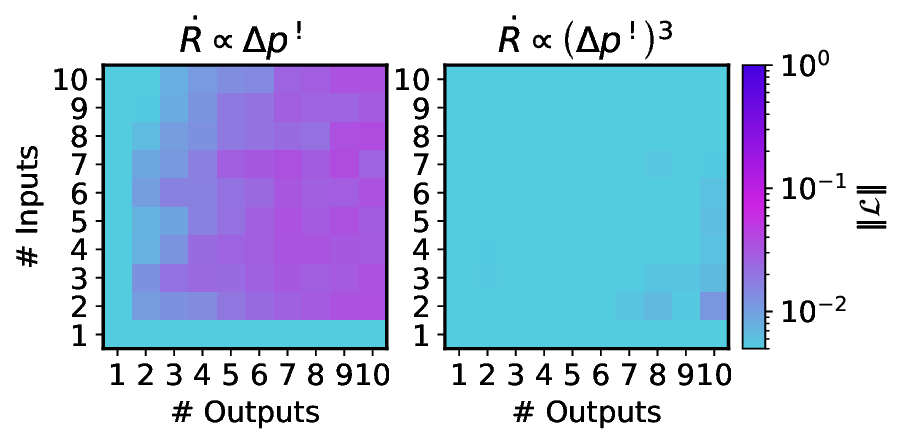}
    }
    \caption{MSE Loss at the end of training for tasks with varying number of inputs and outputs. In the left panel the local rule of $\dot{R} \propto   \Delta p^!$ was used and in the right $\dot{R} \propto \left(\Delta p^!\right)^3$. The loss shown (ranging in color from cyan to purple) is an average over $8$ runs with randomly chosen $M$, and over the final $400$ time steps in each run. During each time step, the loss in normalized by the mean squared value of the desired outputs to result in a dimensionless quantity. BEASTS with the non-linear evolution rule perform better than the linear rule in large scale tasks with multiple inputs and outputs. For specific parameters that were used, the reader is referred to appendix \ref{app:net_structure}.}
    \label{fig:log_loss_afo_inputs_outputs}
    \end{figure}

\subsection{Linear Regression -  non-linear evolution rule}\label{sec:scaling_up}
    
    To address the decrease in performance as task complexity increases when using a linear evolution rule,
    we examine a non-linear evolution rule:

    \begin{equation}\label{eq:Rdot_afo_deltap_nonlin}
        \dot{R}_{ij}\left(t\right)=\gamma  \left(\Delta p^!_{ij}\left(t\right)\right)^3 \ ,
    \end{equation} 
    where changes in resistance emphasize edges where the pressure drops are larger.
    Changes in resistance decrease naturally in time as the loss decreases while using the linear rule (figure \ref{fig:performance_2_networks}), but they decrease much more rapidly using the non-linear rule, (equations \ref{eq:Adaline-like} and \ref{eq:Rdot_afo_deltap_nonlin}), so training halts before completion. To prevent the slowdown, $U^\dagger\overrightarrow{\Delta p}$ is normalized to a magnitude of $1$, and the learning rate $\alpha$ is decreased exponentially over time (Appendix \ref{app:net_structure}). Then the system could settle in a loss minimum. This is a deterministic annealing procedure that negates slowdowns.
    The non-linear local rule shows exceptional learning capabilities, qualitatively better than the linear local evolution rule, as shown in the right panel of figure \ref{fig:log_loss_afo_inputs_outputs}. 
    
    We examined whether the improvement is due to the non-linear local rule being more aligned with GD than the linear rule,  
    by measuring the cosine similarity between the linear rule and GD and the non-linear and GD. None seemed to be significantly higher than the other (Appendix \ref{app:cosine_lin_nonlin}). Moreover, in Appendix \ref{app:comparison_GD} we show that even though BEASTAL yields different sets of DOFs than if the network was trained using exact GD, the internal DOFs change mostly down the gradient of the loss while using BEASTAL.
    
    We suggest a heuristic explanation of why the non-linear evolution rule performs better than the linear one: a single-layer fully-connected network can solve linear regression tasks with arbitrarily low loss. The difficulty is to find a set of internal DOFs that produces the solution, since BEASTAL changes only $\#_{\text{Inputs}}+\#_{\text{Outputs}}$. 
    Hence, BEASTAL can evolve the system along limited directions in the full loss landscape. From equations \ref{eq:Rdot_afo_deltap} and \ref{eq:Adaline-like}, the evolution according to the linear local rule is $\dot{R}_{ij}\propto U^\dagger\left(y_{i} - x_{j}\right) \left(\hat{y}_{i}-y_{i}\right)$ which is linear in the loss. All resistances change at each time step, as long as the loss is non-zero, and resistances keep evolving. 
    However, the loss plateaus after some time, while the resistances keep changing. Thus, some resistances change in a direction that lowers the loss while others change in a direction that increases it, and the contributions cancel out on average. Using the non-linear evolution rule, $\dot{R}_{ij}\propto U^\dagger\left(y_{i} - x_{j}\right)^3 \left(\hat{y}_{i}-y_{i}\right)^3$, greatly discriminates for changes where the loss is significant. This is similar to changing only the resistances that most influence the loss at each step, and the system eventually finds a set of resistances that solve the task.

\subsection{Classification}\label{sec:Classification}

\begin{figure}[ht]
      \centering
      \setlength{\unitlength}{1mm}  
      \begin{overpic}[width=0.9\linewidth, grid=false]{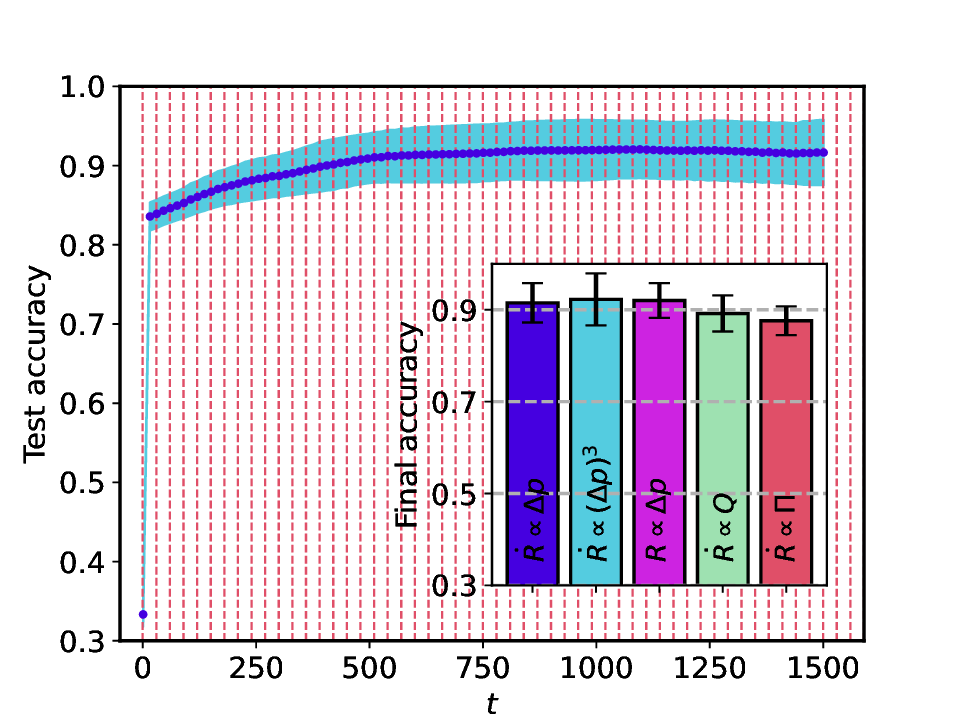}
        \put(17,20){\includegraphics[width=34mm]{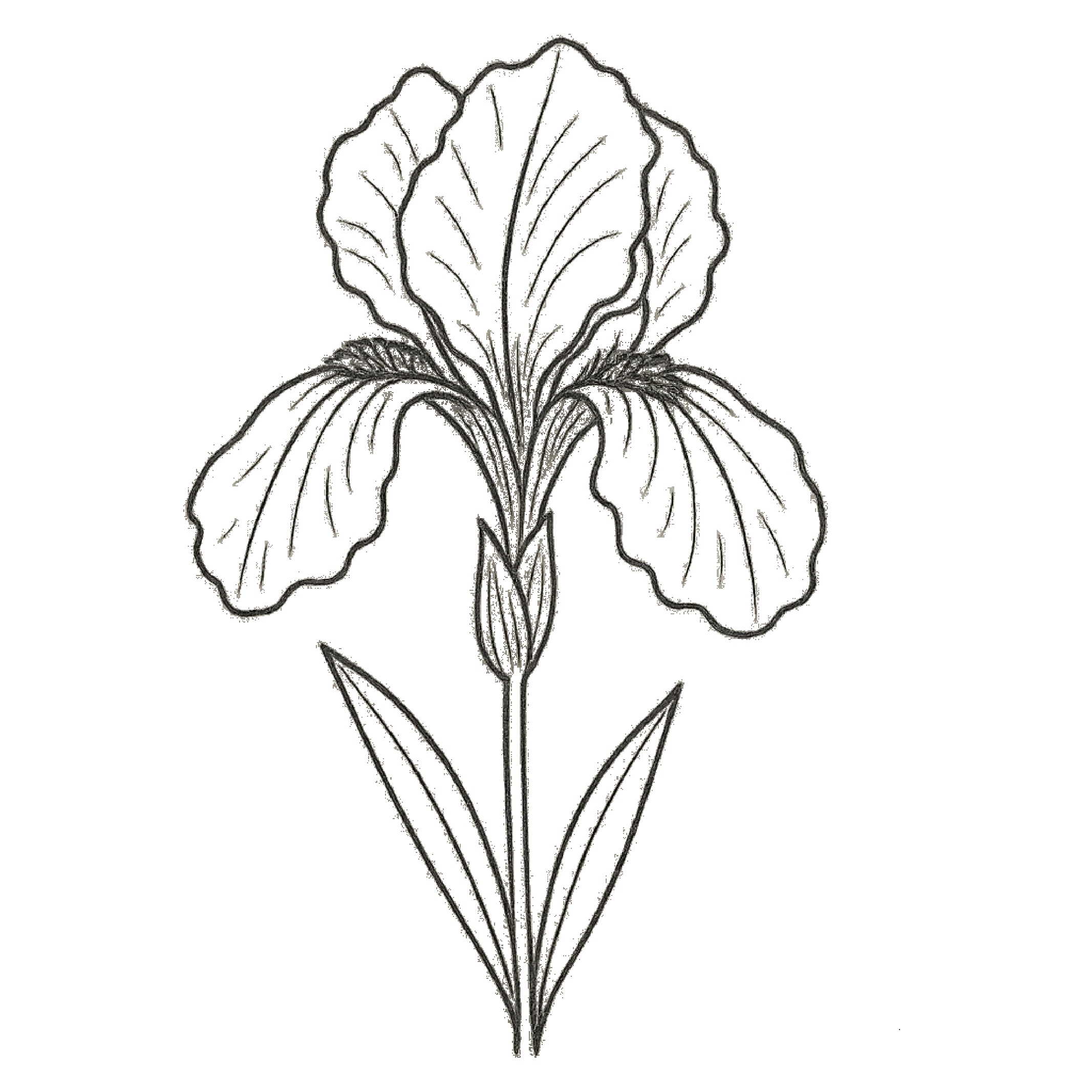}} 
      \end{overpic}
      \caption{Test accuracy as a function of training step 
              $t$, defined as the fraction of irise flowers correctly classified from the test set. Purple dots indicate the ensemble mean over 8 runs, with shaded regions representing one standard deviation. Red dashed lines mark steps at which the desired outputs were updated (every 30 steps). The inset shows final test accuracy for networks with different local evolution rules, trained on the Iris dataset. Here $5$ BEASTS were studied, relating to $5$ different physical materials: $\dot{R}_{ij}=\gamma \Delta p^!_{ij},\,\dot{R}_{ij}=\gamma \left(\Delta p^!_{ij}\right)^3,\, \dot{R}_{ij}=\gamma Q^!_{ij},\,\dot{R}_{ij}= \gamma \Pi^!_{ij}$ and one where the resistance itself is a function of the instantaneous pressure drop $R_{ij}\left(t\right)=\gamma \Delta p^!_{ij}\left(t\right)$. The resulting test accuracy is averaged over $8$ runs and error bars are a standard deviation from average. All networks show the ability to learn reaching accuracy of above $85\%$, on average, within less than $2000$ time steps.}
      \label{fig:accuracy_classification}
    \end{figure}

    BEASTAL is tested on the iris dataset \cite{fisher1936use}, where one of $3$ iris speciess is classified according to $4$ input values (or "attributes").
    The iris dataset is a benchmark for classification tasks; linear networks trained using logistic regression can accurately classify up to $98\%$ of the iris flowers. Breaking the ceiling of $98\%$ requires non-linear networks which are outside the scope of this work. Classification, regardless of network linearity, is a staple for network generalizability since the network learns an abstract boundary between classes whereas linear regression requires a direct linear relation between inputs and outputs.
    We tokenize the iris labels as $3$ dimensional output (see appendix \ref{app:net_structure}), so a correct classification of an iris sample is when the measured output is closest to the desired output of that sample, based on the $L2$ norm. The test accuracy is the fraction of samples from a test set that were correctly classified (for more information see Appendix \ref{app:classification}).    
    Using the BEASTAL scheme, we used a training set of $30$ Iris samples and the remaining $120$ were reserved for testing. Test accuracy exceeds $93\%$ on average after a few epochs, as shown in figure \ref{fig:accuracy_classification}.

\section*{Conclusions}\label{sec:conclusions}

    BEASTAL schemes are versatile, robust to initial resistance values and numbers of input/output nodes, and the learning is decentralized, i.e. uses local rules, with access only to the system boundaries. Performance can be enhanced by introducing a non-linear local evolution rule. In this case the computation is still linear, and limited in the task that it can perform. We reserve the exploration of nonlinear networks to future study.
    BEASTAL is applicable to a wide variety of physical materials, i.e. with different evolution rules as shown in the inset in figure \ref{fig:accuracy_classification}. Moreover, the same system with a different learning rule has a different set of resistances, even when starting from an identical state. This suggests that each system learns the task in a unique way. Generally, the system evolves to a different state of resistances than the state it evolves to using exact GD. This may be an advantage, if one is able to decipher whether it was trained using the BEASTAL scheme or GD. 
    
    BEASTAL differs from Contrastive Learning, which also trains BEASTS through local evolution, in that 
    it does not contrast two different states of the system using 
    engineered learning rules.
    BEASTAL is, in essence, a form of "exaggerated aging": the imposed inputs and outputs are amplified in the directions that lower the loss. Note that unlike Directed Aging, the imposed values do not take the form of the desired task itself.    
    
    BEASTS are modeled in this work as fluidic resistor networks at steady state but the framework applies to any physical network that minimizes power dissipation or energy \cite{berneman2024designing, marbach2021network}. 
    Physical implementations on common materials could, for example, be electrical resistor networks where resistances increase as they heat up due to local power dissipation \cite{horowitz1989art}, spring networks where springs weaken or strengthen as they are stretched and heated \cite{buckner2019enhanced}, or fluidic networks where channels clog and unclog \cite{majekodunmi2022flow}. In the latter, for example, the Measurement modality can be performed over short timescales, during which elements do not have time to change their resistances \cite{falk2025temporal, liu2015spiking, yan2016neuromorphic}.



\appendix

\renewcommand{\thefigure}{A\arabic{figure}}
\setcounter{figure}{0}  

\section{Adding hidden layers}\label{app:hidden_layers}

\begin{figure}[ht]
\centerline{   \includegraphics[width=\columnwidth,keepaspectratio]{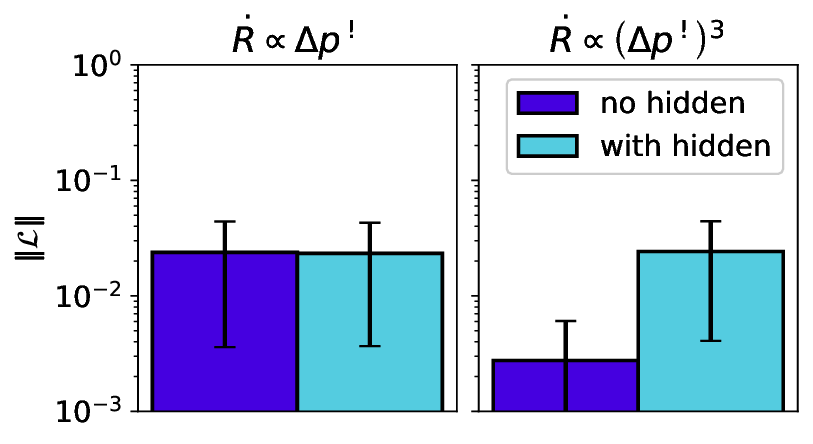}
    }
    \caption{Comparison of the absolute value of the loss at the end of training using the BEASTAL scheme for networks with and without a hidden layer, for a typical example of $6$ input $7$ output task. In the left panel: networks with a linear local rule and in the right: networks with a non-linear rule. Each bar is a single task and error bars are a standard deviation of the loss from the mean at the final $100$ time steps during the run. The scheme performs well even in networks with a hidden layer, but the best performance is achieved using a single layer fully connected network with a non-linear local rule.}
    \label{fig:comparison_hidden_layers}
    \end{figure}

We examined networks with a hidden layer of nodes, in addition to the inputs, outputs and ground. All inputs are fully connected to all the nodes in the hidden layer, and the hidden layer is fully connected to the outputs. The number of nodes in the hidden layer is $\max\left(\#_{\text{inputs}},\#_{\text{outputs}}\right)$. Examining different task matrices $M$, local rules and number of inputs and outputs, we see that networks with a hidden layer perform equally well as those without, as long as the number of inputs or the outputs is $1$; Tasks with only a single input or output can be solved to arbitrarily low loss, as was shown in figure \ref{fig:log_loss_afo_inputs_outputs}. This holds for both linear and non-linear evolution rules and for networks with and without a hidden layer.
On the other hand, for tasks where $\left(\#_{\text{inputs}}+\#_{\text{outputs}}\right)<\left(\#_{\text{inputs}}\times\#_{\text{outputs}}\right)$, networks that a) do not have a hidden layer and b) have a non-linear local rule, perform best. Only then, the loss is lowered to values below $10^{-2}$, as seen in Figure \ref{fig:comparison_hidden_layers}. Networks with a hidden layer perform as good as a single layer network with a linear rule. This suggests a limit to the applicability of the BEASTAL scheme to networks with hidden layers; it performs well only on tasks where there is only a single input or output. Otherwise, a single fully connected network with a non-linear evolution rule is preferred.

\section{Derivation of the BEASTAL scheme from gradient descent and the Adaline algorithm}\label{app:derivation_Adaline_like}

A common method for training an artificial neural network is to evolve its weights $W_{ij}$ down the gradient of a cost function $c$, as $W_{ij}\left(t\right)=W_{ij}\left(t-1\right)-\alpha \frac{\partial c}{\partial W_{ij}}$. In the BEASTS studied here, there are no weights per se. There is, however, a way to present the outputs as a multiplication of the inputs by an effective weights matrix, if considering the conductances $k_{ij}=1/R_{ij}$; for bipartite, fully connected, single layer network where the outputs are connected to a ground node, the network outputs given the drawn inputs $\vec{x}$ are:

\begin{equation}\label{eq:network_output}
    y_{i}=\frac{\sum_{j}k_{ij}x_{j}}{\sum_{l}k_{il}} \ ,
\end{equation}

where $j$ goes over all input nodes and $l$ over all inputs and the ground node.
In BEASTS, the BEASTAL scheme is an online training algorithm where at each training step an input $\vec{x}$ is drawn, the desired output $\hat{\vec{y}}$ is compared to the measured $\vec{y}$ and the resistances / conductances change incrementally.
For a regression task, i.e. $\hat{\vec{y}}=M\vec{x}$, the gradient of the MSE loss 
$c=\frac{1}{N_{out}}\left|\hat{\vec{y}}-\vec{y}\right|^{2}$ with respect to conductances for every drawn input is

\begin{equation}
    \frac{\partial c}{\partial k_{ij}}=\frac{2}{N_{out}}\left(\frac{\sum_{l}k_{il}x_{l}}{\left(\sum_{l}k_{il}\right)^{2}}-\frac{x_{j}}{\sum_{l}k_{il}}\right)\left(\hat{y}_i-y_i\right) \ ,
\end{equation}

and noting that the middle bracket term is just $y_i - x_j$ divided by the sum of conductances, 

\begin{equation}
    \frac{\partial c}{\partial k_{ij}}=\frac{2}{N_{out}}\left(y_{i}-x_{j}\right)\left(\hat{y}_{i}-y_{i}\right) \frac{1}{\sum_{l}k_{il}} \ .
\end{equation}

The term proportional to conductances does not depend on the specific drawn input or measured output, only on the state of the fluidic resistor network. Therefore, it is changing slowly and could be considered as yet another learning rate factor. The equation then boils down to

\begin{equation}\label{eq:Adaline-like-cost}
    \frac{\partial c}{\partial k_{ij}} \propto \left(y_{i}-x_{j}\right) \left(\hat{y}_{i}-y_{i}\right) \ .
\end{equation}

Considering conductances as weights, equation \ref{eq:Adaline-like-cost} is strikingly similar to the Adaline algorithm; the ADAptive LInear NEuron algorithm trains a single layer, bipartite, fully connected artificial neural network with a linear activation and MSE cost. There, the weights are changed according to $\dot{W}_{ij}=-\frac{\partial c}{\partial W_{ij}}=\alpha x_j \left(\hat{y}_i-y_i\right)$ \cite{widrow1988adaptive}. This rule is proven to evolve the weights down the instantaneous gradient of the MSE cost at every training sample. Under slow changes in resistance, $\dot{k}_{ij}=-\dot{R}_{ij}$, so evolving the resistances down the gradient of the loss takes the form 

\begin{equation}\label{eq:Adaline-implementation_resnets}
    x_j^!-y_i^!=\alpha \left(y_{i}-x_{j}\right)\left(\hat{y}_{i}-y_{i}\right) \ .
\end{equation}

This is an Adaline algorithm implementation on fluidic resistor networks with the local rule as in equation \ref{eq:Rdot_afo_deltap}. Moreover, it involves only given inputs and measured or desired outputs. It is, however not directly applicable to BEASTS, since it requires direct access to all internal parameters; this a set of $\left(\#_{\text{inputs}}+1\right)\times \#_{\text{outputs}}$ equations for pressure differences but there are no $\#_{\text{inputs}}+\#_{\text{outputs}}$ Update modality pressures to satisfy them all. 

The closest version of the Adaline algorithm for BEASTS requires taking the pressure differences in \ref{eq:Adaline-implementation_resnets} in vector form, and multiplying it vector by the pseudo-inverse of the incidence matrix, as introduced in Appendix \ref{app:incidence_mat}. This gives the values for inputs and outputs during the Update modality:

\begin{equation}
    \left[\begin{array}{c}
    \vec{x}^{!}\\
    \vec{y}^{!}
    \end{array}\right]=\frac{\alpha}{\gamma}U^{\dagger}\overrightarrow{\Delta p^!}
\end{equation}

with $\overrightarrow{\Delta p^!}=\overrightarrow{\left(x^!_j-y^!_i\right)}=\overrightarrow{ \left(y_{i} - x_{j}\right) \left(\hat{y}_{i}-y_{i}\right)}$.

$\left[\begin{array}{c}
    \vec{x}^{!}\\
    \vec{y}^{!}
\end{array}\right]$ is a vector with dimensions $\#_{\text{Inputs}} \times\#_{\text{Outputs}}$ containing values to impose during the Update modality after each Measurement, relying solely on the previous Measurement and the loss. 

The Update modality values for system with a non-linear local rule, as mentioned in section \ref{sec:scaling_up}, is slightly altered to an annealing of the learning rate:

\begin{equation}
    \left[\begin{array}{c}
    \vec{x}^{!}\\
    \vec{y}^{!}
    \end{array}\right]=\frac{\alpha\left(t\right)}{\gamma}U^{\dagger}\frac{\overrightarrow{\Delta p^!}}{\|\overrightarrow{\Delta p^!}\|}
\end{equation}
where $\alpha\left(t\right)=\alpha_0\exp\left(-\beta t/T\right)$ and $T$ is the total training time.

\section{Incidence matrix}\label{app:incidence_mat}

$U$ is the incidence matrix relating edges to nodes \cite{rocks2017designing, rocks2019limits}, whose dimensions are $\#_{\text{edges}}\times\#_{\text{nodes}}$; each row corresponds to an edge and each column corresponds to a node. Each row there has an entry of $1$ at the index of an incoming node and $-1$ at the outgoing. However, we stress that the network is not a directional network and the order of the positive or negative $1$ does not matter. For a $2$ inputs $2$ outputs and $1$ ground node bipaprtite fully connected network where outputs are connected to ground, $U$ is

\begin{equation}
    U=\left[\begin{array}{ccccc}
    1 & 0 & -1 & 0 & 0\\
    0 & 1 & -1 & 0 & 0\\
    1 & 0 & 0 & -1 & 0\\
    0 & 1 & 0 & -1 & 0\\
    0 & 0 & 1 & 0 & -1\\
    0 & 0 & 0 & 1 & -1
    \end{array}\right]    
\end{equation}

Where the inputs are at columns $1,\,2$, outputs are at $3,\,4$ and the ground is in the last column.
From there, equation \ref{eq:Rdot_afo_deltap} could be written as $\dot{\vec{R}}=\gamma U \vec{p}^{\,!}$ where $\vec{p}^{\,!}$ is a vector of the pressures of length $\#_{\text{nodes}}$.

\section{Comparison to stochastic gradient descent}\label{app:comparison_GD}

We compare the BEASTAL scheme to standard Gradient Descent (GD) in which learning DOFs evolve in time down the gradient of a cost function $c$. Here, the learning DOFs are conductances $k_{ij}=\frac{1}{R_{ij}}$ and the evolution rule takes the form $k_{ij}\left(t\right)=k_{ij}\left(t-1\right)-\alpha \frac{\partial c}{\partial k_{ij}}$, as in \cite{stern2021supervised}. 
This evolution rule is not a local physical one and as a means of comparison we implement it by directly changing the value of the conductances, disregarding equation \ref{eq:Rdot_afo_deltap}. We again choose the MSE as the cost function, $c=\overline{\|\hat{\vec{y}}-\vec{y}\|^2}$, at each training step, to conform to the BEASTAL scheme. We stress that since every training step involves a single training sample, it is in essence a stochastic GD with batch size of $1$.

\begin{figure}[ht]
\centerline{   \includegraphics[width=\columnwidth,keepaspectratio]{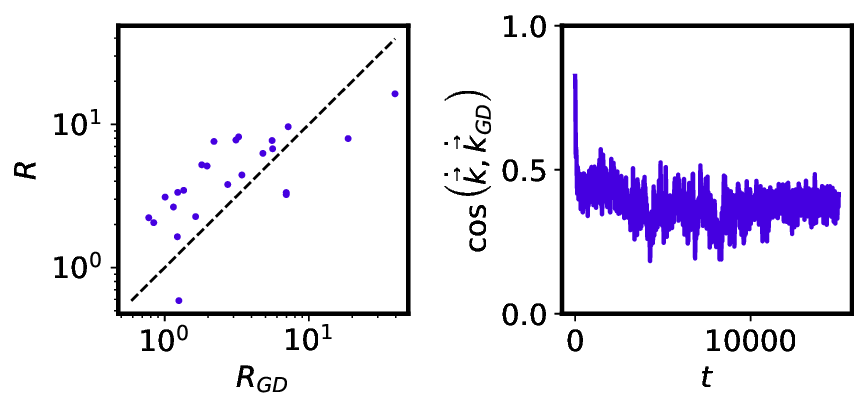}
    }
    \caption{Comparison between BEASTAL and the gradient descent algorithm. A $4$ input $6$ output network was trained on a regression task using  both schemes. Left panel: edge resistances at the end of training using the Adalike algorithm versus the corresponding edge resistance using GD. The final resistances rarely sit on the diagonal, meaning that the schemes end up with different learning DOFs at the end of training. 
    Right panel: Cosine similarity between changes in conductances $\dot{\vec{k}}$ using the BEASTAL scheme and those calculated using GD for every training sample when training is done using BEASTAL. The similarity is positive meaning generally the system evolves mostly down the gradient of the loss}
    \label{fig:comparison_GD}
    \end{figure}

First, training the network using BEASTAL results in different resistances at the end of training than proper GD, as shown in the left panel of figure \ref{fig:comparison_GD}.
Second, from equation \ref{eq:network_output}, a network that solves the task $y_i=M_{ij}x_j$ has $\frac{\frac{1}{R_{ij}}}{\Sigma_l \frac{1}{R_{lj}}}=M_{ij}$ where $l$ goes over all nodes that connect to output $j$. This means that two different networks that solve the task have the same resistances leading to output node $j$, up to multiplication by a constant $\Sigma_l \frac{1}{R_{lj}}$. 
Resistances are therefore arranged in $6$ series of $4$ in figure \ref{fig:comparison_GD}, since there are $6$ outputs and $4$ inputs. The two networks only differ in $6$ constants from each other. For networks with $6$ outputs, there are $6$ constants that can be changed without changing the output, so the manifold of zero loss is $6$ dimensional and we see that BEASTAL and GD reach different points in this manifold. Additionally, the non-linear scheme tends to produce higher resistances than GD, which is left for future studies. 
Lastly, the evolution of the system under the BEASTAL scheme is largely in the same direction as if GD was used: the cosine similarity between the direction of change in conductance using the BEASTAL scheme and $-\frac{\partial c}{\partial k}$, both calculated during the evolution of the network that used BEASTAL, is positive (right panel). In a high dimensional resistance space, even a small positive cosine similarity means a significant correlation between the vectors. Here the resistance space is $24$ dimensional and the cosine similarity approaches $0.5$ so the evolution is really mostly down the gradient of the loss. We emphasize that GD is not physically applicable to BEASTS since it requires computation of- and access to- all learning DOFs. The reader is referred to Appendix \ref{app:net_structure} for network parameters. We conclude that even though the BEASTAL scheme largely follows GD, it yields a different set of resistances that solve the task than if GD was used.

\section{Parameters and network structure}\label{app:net_structure}

In a linear regression task performed using fluidic resistor networks, $M$ cannot obtain any arbitrary value since for positive resistances the outputs cannot have higher values than the inputs; it is constrained so that the entries of each line in $M$ sum up to less than $1$. However, if a general $M$ is desired, dividing the whole task matrix $M$ by a constant and multiplying the outputs by the same constant will yield the desired task. 

For a regression task where the task matrix $M$ is $\#_{\text{Outputs}}\times\#_{\text{Inputs}}$ there are $\#_{\text{Outputs}}$ equations for $\#_{\text{Inputs}}$ parameters. The structure of all networks (except for Appendix \ref{app:hidden_layers}) was chosen as the following: $\#_{\text{Inputs}}$ input and $\#_{\text{Outputs}}$ output nodes, bipartite fully connected, and each output was connected to a single ground node. Hence, the networks consisted of $\#_{\text{Inputs}}+\#_{\text{Outputs}}+1$ nodes and $\#_{\text{Inputs}}\times \#_{\text{Outputs}}+\#_{\text{Outputs}}$ edges (resistors) in total. 
The choice of task matrix $M$ was from a uniform random distribution, but such that every line of $M$ sums up to $0.75$ at most. This ensures that the desired outputs are sufficiently low; the outputs cannot exceed the value of the maximal input. 

For the Iris classification task, networks with $4$ input and $3$ output nodes that are bipartite fully connected were used, and each was connected to a single ground node. The training set consisted of $30$ randomly chosen Iris samples, the remaining $120$ left for test, under which the test accuracy was measured, and the target outputs were re-calculated as in section \ref{sec:results} at the beginning of every epoch, so every $30$ training steps (see Appendix \ref{app:classification} for further detail). All along the work $\gamma=1$ was used, and $\alpha$ in the range $\left[0.01-1.5\right]$. For the annealing procedure where local rule is non-linear (Appendix \ref{app:derivation_Adaline_like}), $\beta$ was in the range $\left[0.5,2\right]$


\section{Scaling up the task - comparison to a non-iterative analytical scheme}\label{app:pseudo_inverse}

For a linear evolution rule, performance decreases as task complexity increases. We have shown that the nonlinear evolution rule successfully resolves the degradation, but here we show that even BEASTS with a linear evolution rule show promising results if trained with BEASTAL; They consistently outperforms a non-iterative analytical scheme for tuning BEASTS. We first present the non-iterative scheme, which deterministically finds the Update modality values for a BEAST with the local rule $R_{ij}=\gamma\Delta p_{ij}$, without learning.
    
In the non-iterative scheme, the supervisor knows the task matrix $M$ itself beforehand, not mere instances of a dataset. The supervisor starts by finding the optimal resistances that could solve the task using equations \ref{eq:power_dissipation} and \ref{eq:task} (regardless of the evolution rule). This requires some external numerical methods which are not within the scope of this paper. The optimal set of resistances is termed $\vec{R}^*$, and is a vector of length $\#_{\text{edges}}$; for a single-layer, fully-connected network and an $M$ matrix whose rows sum up to less than $1$, such a set is guaranteed to exist. In fact, for a network with $\left(\#_{\text{inputs}}+1\right)\times \#_{\text{outputs}}$ edges, there is a $\#_{\text{outputs}}$ dimensional manifold where resistances solve the task exactly and $\vec{R}^*$ lies on that manifold. Here, the supervisor aims to tune the Update values to yield $\vec{R}^*$, even though the specific $\vec{R}^*$ often can't be directly achieved by the BEAST.
Note that $R_{ij}=\gamma\Delta p^{!}_{ij}$ could be written as $\vec{R}=\gamma U \vec{p}^{\,!}$ where $\vec{p}^{\,!}$ is a vector of length $\#_{\text{nodes}}$ and $U$ is the incidence matrix.  
From there, the supervisor imposes the Update modality input and output pressures as $\vec{p}^{\,!}=\frac{1}{\gamma}U^{\dagger}\vec{R}^*$. These update modality values yield resistances $\vec{R}=UU^{\dagger}\vec{R}^*$; in the unique case where $U$ is invertible, $\vec{R}=\vec{R}^*$, but for tasks where there is more than $1$ input and $1$ output, $U$ is non-invertible. These resistances are the realizable resistances which are closest to $\vec{R}^*$ in the examined BEAST. 
    
This non-iterative scheme performs worse than a network trained using the BEASTAL scheme, in all examples tested where the number of inputs and outputs exceeded $1$. Specifically, for a typical example of a $3$ input $2$ output task, the non-iterative scheme reached an average normalized MSE loss of 
$\overline{\|\vec{\mathcal{L}}\|^2}=1.2$ while the iterative training scheme reached a significantly lower value of $\overline{\|\vec{\mathcal{L}}\|^2}=5\times 10^{-2}$ for the linear rule and $\overline{\|\vec{\mathcal{L}}\|^2}=1\times 10^{-6}$ for the non-linear rule. 

The resistances found using BEASTAL are different than those found using the non-iterative scheme. Even when initialized with the non-iterative scheme’s resistance configuration, BEASTAL finds alternative resistances, based on incremental feedback, allowing the network to discover configurations that function more effectively.
This suggests that attempting to tune BEASTS by calculating the optimal values for its learning DOFs as if they are realizable is not a good strategy, and even the knowledge of what the optimal values are is not useful. It is better to train the system iteratively while presenting it with  samples of the task regardless of the optimal learning DOFs are. This can be attributed to the fact that the non-iterative scheme, which relies on the pseudo-inverse of the incidence matrix connecting edges and nodes, attempts to set internal degrees of freedom directly to optimal values, even when these values cannot be physically realized without direct access. In contrast, the BEASTAL scheme complies with the physical constraints of the system, evolving the resistances through iterative adaptation using samples of inputs and desired outputs and minimizing a loss. The iterative mechanism yields better performance, even though the resulting resistances deviate from theoretical optimal values, since the former are inaccessible for BEASTS.

\section{Cosine similarity comparison - linear and non-linear schemes}\label{app:cosine_lin_nonlin}

\begin{figure}[ht]
\centerline{\includegraphics[width=\columnwidth,keepaspectratio]{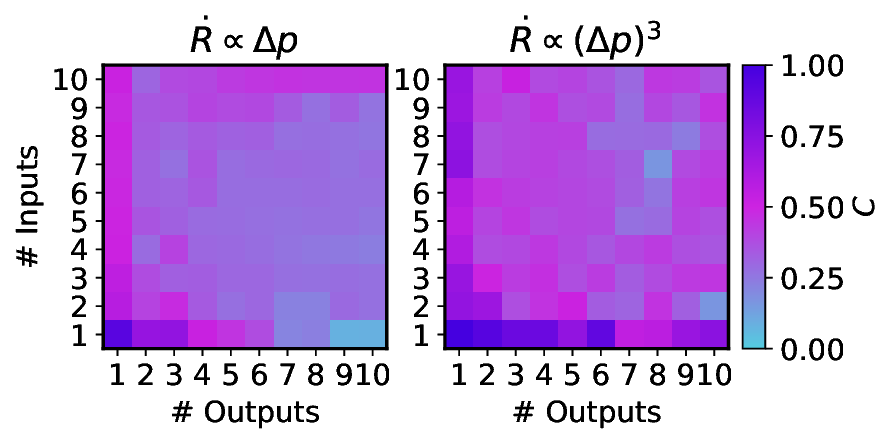}}
    \caption{The average cosine similarity $C$ between the linear rule and GD (left panel) and the non-linear rule and GD (right panel), as function of number of inputs and outputs. The average is over $4$ runs with randomly chosen $M$, and over the whole training time in each run.}
    \label{fig:cos_sim_lin_nonlin}
    \end{figure}

We examined whether the improvement is due to the non-linear local rule being more aligned with GD than the linear rule. A common method to study how much the state of a system aligns with GD is to measure the cosine similarity between the vector of change in DOFs $\overrightarrow{\dot{k}_{ij}}$ and the calculated gradient of the loss $-\overrightarrow{\frac{\partial\text{Loss}}{\partial k_{ij}}}$, as in $C=-\frac{\overrightarrow{\dot{k}_{ij}}\,\cdot\,\overrightarrow{\frac{\partial\text{Loss}}{\partial k_{ij}}}}{\|{\overrightarrow{\dot{k}_{ij}}}\|\|{\overrightarrow{\frac{\partial\text{Loss}}{\partial k_{ij}}}}\|}$. We measured $C$ for the non-linear and linear rules; None seemed to be significantly higher than the other (figure \ref{fig:cos_sim_lin_nonlin}).

\section{Classification task - tokenizing the output vector and additional information}\label{app:classification}

Typically, each species would be represented by a unique three-dimensional output vector with a single active entry (often called a 'one-hot label' vector) for classification. Instead, the average output vector of all samples from each species was used.
This requires passing all $150$ iris samples in the dataset through the network at the beginning of each epoch and averaging over outputs from all samples of the same species. 
Nonetheless, since the network is linear, this averaging is equivalent to passing the average of the inputs of that species, which simplifies the training scheme; it is essentially tokenizing the outputs to an embedding dimension of $3$, a common practice in training neural networks. 
It is worth noting that the desired outputs are determined by the network parameters, since they are an outcome of passing all of the input dataset. Since the internal parameters change in the course of training, the desired outputs change as well; nonetheless, the three vectors of desired output (one for each speciess) reach steady-state after a few epochs of training.

\nocite{*}
\bibliography{citations}

\end{document}